\documentclass[lettersize,journal]{IEEEtran}
\usepackage{amsmath,amsfonts}
\usepackage{algorithmic}
\usepackage{array}
\usepackage{float}
\usepackage{textcomp}
\usepackage{stfloats}
\usepackage{url}
\usepackage{verbatim}
\usepackage{graphicx}
\usepackage{subfigure}
\usepackage{booktabs}
\usepackage{subcaption}
\usepackage{algorithm}
\usepackage{enumitem}
\usepackage{multirow} 
\def\BibTeX{{\rm B\kern-.05em{\sc i\kern-.025em b}\kern-.08em
    T\kern-.1667em\lower.7ex\hbox{E}\kern-.125emX}}
\usepackage{balance}
\IEEEoverridecommandlockouts
\begin{document}
\title{Balanced Edge Pruning for Graph Anomaly Detection with Noisy Labels}

\author{
\IEEEauthorblockN{Zhu Wang\IEEEauthorrefmark{1},
Junnan Dong\IEEEauthorrefmark{1},
Shuang Zhou\IEEEauthorrefmark{1},
Chang Yang\IEEEauthorrefmark{1},
Shengjie Zhao \IEEEauthorrefmark{2}\thanks{Corresponding author: Shengjie Zhao, Email: shengjiezhao@tongji.edu.cn}, 
Xiao Huang\IEEEauthorrefmark{1}
}

\IEEEauthorblockA{\IEEEauthorrefmark{1}The Hong Kong Polytechnic University, Hung Hom, Hong Kong SAR \\
Email: \{juliazhu.wang, hanson.dong, shuang.zhou,  chang.yang\}@connect.polyu.hk,  xiaohuang@polyu.edu.cn} 

\IEEEauthorblockA{\IEEEauthorrefmark{2}Tongji University, Shanghai, China \\
Email: shengjiezhao@tongji.edu.cn
}
}

\markboth{Journal of \LaTeX\ Class Files,~Vol.~18, No.~9, September~2020}%
{How to Use the IEEEtran \LaTeX \ Templates}

\maketitle

\begin{abstract}
Graph anomaly detection (GAD) is widely applied in many areas, such as financial fraud detection and social spammer detection. Anomalous nodes in the graph not only impact their own communities but also create a ripple effect on neighbors throughout the graph structure. Detecting anomalous nodes in complex graphs has been a challenging task. While existing GAD methods assume all labels are correct, real-world scenarios often involve inaccurate annotations. These noisy labels can severely degrade GAD performance because, with anomalies representing a minority class, even a small number of mislabeled instances can disproportionately interfere with detection models. Cutting edges to mitigate the negative effects of noisy labels is a good option; however, it has both positive and negative influences and also presents an issue of weak supervision. To perform effective GAD with noisy labels, we propose \textbf{RE}inforced \textbf{G}raph \textbf{A}nomaly \textbf{D}etector (REGAD) by pruning the edges of candidate nodes potentially with mistaken labels. Moreover, we design the performance feedback based on strategically crafted confident labels to guide the cutting process, ensuring optimal results. Specifically, REGAD contains two novel components. (i) A tailored policy network, which involves two-step actions to remove negative effect propagation step by step. (ii) A policy-in-the-loop mechanism to identify suitable edge removal strategies that control the propagation of noise on the graph and estimate the updated structure to obtain reliable pseudo labels iteratively. Experiments on three real-world datasets demonstrate that REGAD outperforms all baselines under different noisy ratios. 

\end{abstract}


\maketitle
\section{Introduction}
Graph anomaly detection (GAD) has emerged as a crucial technique in various real-world scenarios, such as fake news detection~\cite{wang_weak_2020,xie_label_2022,cheng_Causal_2021}, fraud detection~\cite{zhang_fraud_2021,dou_enhancing_2020}, and social spammer detection~\cite{li_RelevanceAware_2021}. This technique can identify outliers that deviate from the majority of normal nodes or communities, which has significant applications for system integrity and security. Moreover, anomalous nodes in graphs can mutually influence normal ones due to their interconnected relations~\cite{ma_comprehensive_2021}, such as robot accounts in social networks. The flexible graph structure has features of large scales and complex networks, making outliers detection valuable and challenging in GAD task~\cite{wang_CrossDomain_2023}. Nevertheless, anomalous nodes not only impact their local communities but can also create a ripple effect on global nodes throughout the graph. Besides, anomalous nodes in graph-based applications can significantly impact across various domains, such as fraudulent accounts in online social networks and irregular transactions in financial networks. 

\begin{figure}[htbp]
    \centering
    \includegraphics[width=0.85\linewidth]{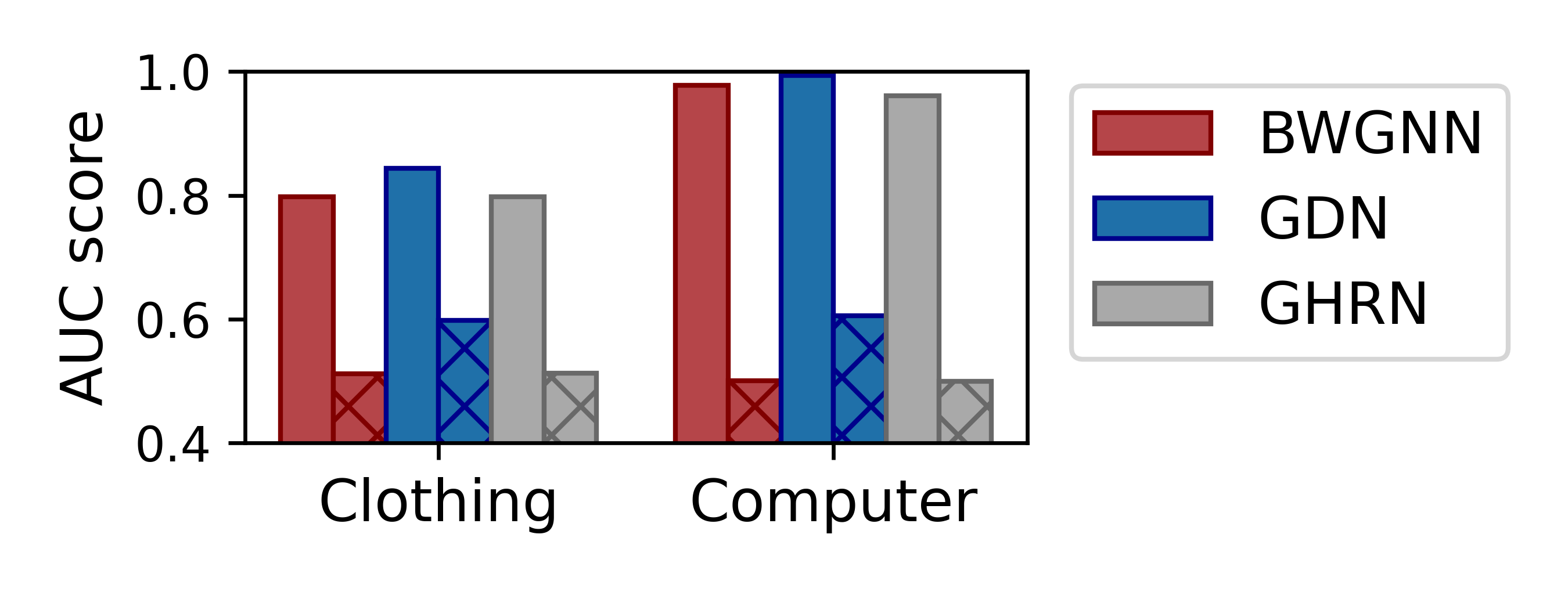}
    \vspace{-8pt}
    \caption{A pilot study reveals that noisy labels degrade the performance of general GAD models. The bars with cross lines denote the results when $0.1\%$ incorrect labels are injected. AUC denotes the metric of the area under the ROC curve.}
    \label{fig:Fig1_pilot}
\end{figure}
Research on graph anomaly detection has gained significant attention and has become an independent topic in recent years. However, traditional GAD methods usually assume that all labels are accurate and focus on the imbalance challenge. Nevertheless, this assumption rarely holds true given currently available annotation techniques, including crowd-sourcing platforms and automated annotation tools~\cite{chen_Entity_2024}, where annotation processes are cheap in cost but obtain low-quality labels. These annotators can be unreliable when labeling anomalous nodes, particularly in large and complex graphs~\cite{dong_ModalityAware_2024}. Anomalies can typically be identified through the analysis of spatial distribution in traditional methods, but the detection becomes challenging when obscured with noisy labels. Noisy labels, even a few, can significantly affect the effectiveness because outliers typically constitute only a small fraction of the overall graph. Nevertheless, a small number of mislabeled instances of anomalous nodes also disproportionately affect the model's ability to identify outliers from numerous normal nodes. In Fig.\ref{fig:Fig1_pilot}, we showcase the impact of noisy labels on two benchmark datasets. We stimulate noise using the label-flipping method in a heuristic way. We observe that noisy labels significantly decline the performance of three semi-supervised detection models. Unfortunately, especially as the scale of the graph expands and the variety of anomalies increases, conventional graph anomaly detection methods under noisy labels struggle to perform well. Thus, it is urgent to develop robust GAD approaches that can maintain high detection accuracy in the presence of noisy labels.


Given the urgent need to develop robust graph anomaly detection approaches, it is important to highlight the challenges posed by noisy labels. Firstly, the presence of noisy labels can confuse the model's ability to identify true anomalous node patterns accurately. Furthermore, anomalous nodes only constitute a small proportion of the graph, and noisy labels exacerbate this imbalance~\cite{liu_pick_2021}. Thirdly, during the data training process, noisy labels result in weak supervision, causing propagating noisy information to neighborhood nodes through complex graph structures~\cite{zheng_AddGraph_2019}. Faced with the above challenges from noisy labels, we leverage reinforcement learning to carefully remove passage transmission from noisy labels to nodes with true labels. Thus, our research problem is pruning suitable edges to improve detection effectiveness under weak supervision. Specifically, there are three problems to be solved: i) how to control the removal of edges properly in a way that reduces interference from incorrect labels in an extremely large space, that is, the entire graph. Cutting edges can bring both positive and negative influences, while it is hard to quantitatively evaluate the consequence of cutting an edge; ii) how to ensure reliable supervision to optimize edge-cutting in the presence of uncertain noisy labels; iii)how to define the feedback to guide the edge-cutting process to control quality and quantity. 
 

To this end, we propose a novel \textit{policy-in-the-loop} framework, the \textbf{RE}inforced \textbf{G}raph \textbf{A}nomaly \textbf{D}etection model (REGAD), aiming at minimizing impacts of noisy labels and enhancing detection effectively. To solve negative effects from wrong-labeled nodes, this model utilizes the edge pruner (a tailored policy network) with an effective reward design by maximizing the performance improvement evaluated by the base detector and uses highly confident pseudo labels to address noisy labels' weak supervision. Specifically, REGAD consists of two sections: a tailored policy network for step-by-step edge pruning and the base detector for delivering trustworthy predictions as pseudo labels, comprised of a policy-in-the-loop mechanism for guiding the cutting strategy exploration. In the first section, we design a two-step action: selecting candidate nodes and then pruning edges centering around the target nodes, a relatively small space. Additionally, we compute rewards by comparing the base detector performance before and after removing edges in actions. The target of the policy network is to maximize the cumulative rewards of all actions. Instead of noisy labels' supervision, we use high-confidence pseudo labels as truth labels, prioritizing cutting edges accurately as expected, which carry interference from candidate nodes possibly with noisy labels. Finally, we design the \textit{policy-in-the-loop} mechanism to iteratively optimize the policy network using the mutual feedback based on reliable pseudo labels and correspondingly evaluate the reconstructed graph structure, targeting to remove the negative information propagation.

In general, we summarize our contributions as below: 
\vspace{-3pt}
\begin{enumerate}
    \item We formally define the problem of noisy label learning for graph anomaly detection.
    \item A tailored policy network is designed to carefully identify noisy labels and optimize the edge pruning by prioritizing the most suspicious nodes as incorrectly labeled outliers or normal nodes. 
    \item We design a novel policy-in-the-loop learning paradigm for GAD with noisy labels. The policy network and the detector complementarily benefit and provide feedback to each other to mitigate the negative impacts of noisy labels.
    \item Extensive experiments are conducted to comprehensively demonstrate the superiority of our framework under different noisy rates on three datasets.
\end{enumerate}
\section{Problem Statement}
\subsection{Notations}
We employ $\mathcal{G}=(\mathcal{V, E},\mathrm{X})$ to denote an attributed graph, where $\mathcal{V}=\left \{  v_{1}, v_{2}, ..., v_{n}  \right \}$ is the set of $n$ nodes, and $\mathcal{E}\subseteq \mathcal{V}\times \mathcal{V} $ is the set of edges. Besides, $\mathrm{X}=\left \{  x_{1}, x_{2}, ..., x_{n}  \right \}$ is the node attributes, $\mathrm{X} \in \mathbb{R}^{n \times d}$, and $d$ is the attribute dimension. $\mathrm{A} \in \mathbb{R}^{n \times n}$ represents the adjacency matrix of $ \mathcal{G}$. If $v_{i}$ and $v_{j}$ are connected, $\mathrm{A}_{ij}=1$. Otherwise, $\mathrm{A}_{ij}=0$. $\mathcal{V}_{L} =\left \{ v_{1}, v_{2},...,v_{l} \right \} $ represents labeled nodes, and  $\mathcal{V}_{U} = \mathcal{V}-\mathcal{V}_{L}$ is the set of unlabeled nodes.  $\mathcal{V}_{L}$ includes normal nodes $\mathcal{V}_{n}$ and abnormal nodes $\mathcal{V}_{a}$. In practice, we only obtain anomaly nodes $\mathcal{V}_{a}$ following $\left | \mathcal{V}_{a} \right| \ll \left | \mathcal{V}_{n} \right|$. $\mathcal{Y}_{L}=\left \{ y_{1}, y_{2}, ..., y_{l} \right \} $ denotes the real ground truth labels of $\mathcal{V}_{L}$. However, in our research problem, we assume that they are not trustworthy and utilize $\bar{\mathcal{Y}}_{L} =\left \{ \bar{y}_{1}, \bar{y}_{2}, ..., \bar{y}_{l}\right \}$ to represent the corrupted ground truth. 

\subsection{Problem Definition}

Given an attributed graph $\mathcal{G}$, the GAD task is formulated as: 
\begin{equation}
    f(\mathcal{G},\mathcal{Y}_{L}) \to \hat{\mathrm{S}},
\end{equation}
where anomaly scores $\hat{\mathrm{S}}$ reflect the likelihood of being an anomalous node. However, under noisy label setting, the ground truth labels $\bar{\mathcal{Y}}_{L}$ are not reliable because a small proportion of labeled nodes $\mathcal{V}_{L}$ are mistaken. Thus, we utilize $\bar{\mathcal{Y}}_{L}$ to represent noisy ground truth labels. For example, the node $v_{i},i\in L$ with the false label $\bar{y}_{i}=0$ is considered as a normal node. Actually, the truth label is $y_{i}=1$, an anomaly node. When we cannot distinguish which are true anomalies and real normal nodes, we aim to predict correct matching anomaly scores $\hat{\mathrm{S}}$ for all nodes by reconstructing the graph structure as much as possible supervised by $\bar{\mathcal{Y}}_{L}$, i.e.,
\begin{equation}\label{eq:detection}
f(\mathcal{G}^{'}, \mathcal{Y}^{'}) \to \hat{\mathrm{S}} ,
\end{equation}
where the detection model $f$ is to estimate the probabilities of nodes being anomalous. In addition, only two label types are considered, anomaly nodes $\mathcal{V}_{a}$ and normal nodes $\mathcal{V}_{n}$ in $\mathcal{V}_{L}$. Nevertheless, labeled nodes are significantly smaller than the number of unlabeled nodes, denoted as $\left | \mathcal{V}_{L} \right| \ll \left | \mathcal{V}_{U} \right|$ in GAD tasks. To stimulate $\bar{\mathcal{Y}}_{L}$, we will leverage the label-flipping introduced in the experiment section. 
\begin{figure*}
    \centering
    \includegraphics[width=1\linewidth,trim = 4.8cm 6.8cm 5.5cm 7.9cm,clip]{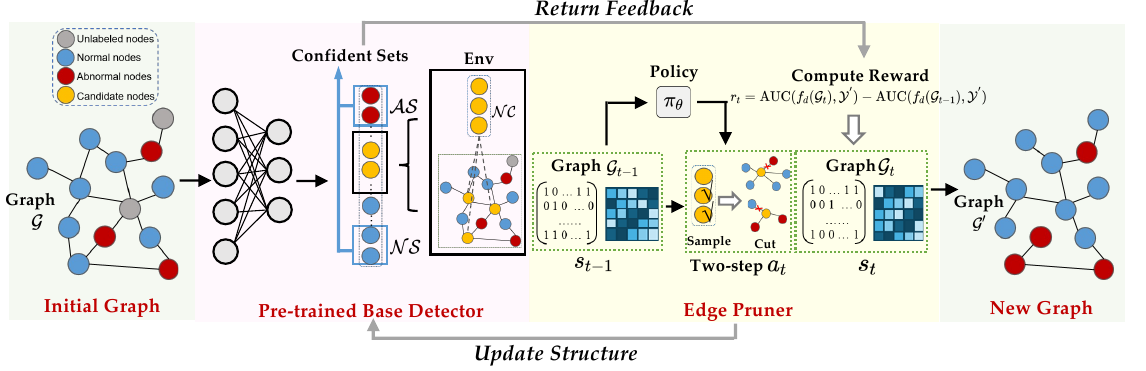}
    \caption{An overview of our framework REGAD with the \textit{policy-in-the-loop} mechanism. The edge pruner explores MDP to obtain a strategy to cut edges and balance the pruned edge quality and quantity by $\pi_{\theta}$ based on the performance improvement. The reward design employs the trustworthy pseudo labels of confident sets from the base detector, i.e., $\mathcal{AS}$ and $\mathcal{NS}$, to guide the reward computation. } 
    \label{fig:frame}
\end{figure*} 
\section{Methodology}
In this section, we propose the \textbf{RE}inforced \textbf{G}raph \textbf{A}nomaly \textbf{D}etection model in Fig.~\ref{fig:frame}. Our model seeks to minimize the impact of nodes with noisy labels on neighborhood nodes by a \textit{policy-in-the-loop} framework with a tailored policy network and a base detector. In particular, we aim to answer three important research questions: i) Given the complicated graph structure, how can we control the number of removed edges in a large search space to control the negative passages from nodes with noisy labels to neighbors? ii) How can we identify potential nodes with false labels and solve the weak supervision from unreliable ground truths? iii) How can we quantitatively evaluate the actions from the edge pruner, i.e., positive (cut accurate edges connecting noisy-label candidates) or negative (miscut edges)? Therefore, we employ the edge pruner (Section~\ref{pruner}) is designed to find a satisfactory policy $\pi_{\theta}(a|s)$, to cut edges between candidates possibly labeled erroneously and nodes with true labels.
Additionally, a base detector (Section~\ref{detector}) provides the basis of reward computation and updates noisy ground truths to improve trustworthiness. The framework (Section~\ref{loop}) evaluates cumulative rewards quantitatively by comparing the performance (AUC) of the base detector before and after graph refinement.

\subsection{Edge Pruner}\label{pruner}

We leverage the edge pruner $f_{e}$, based on the reinforcement learning method, to select candidate nodes and remove suspicious edges of targets in a small search space rather than the entire graph. Our pruner aims to learn a cutting strategy to reshape the graph while keeping pattern learning of anomalous nodes. The refined graph after edge-cutting prevents information transmission from nodes with incorrect labels. Specifically, if the target is a true normal node, edge-cutting hardly has negative effects due to numerous relations connecting with similar normal nodes. In contrast, when it is an outlier, the pruner reduces the assimilation for this node from neighborhood normal nodes but highlights the impact of node features on label prediction.

\begin{figure}[htbp]
    \centering
    \subfigure[Over-cutting]{
        \label{figa}
        \begin{minipage}[b]{0.4\linewidth}
            \centering
            \includegraphics[scale=0.4]{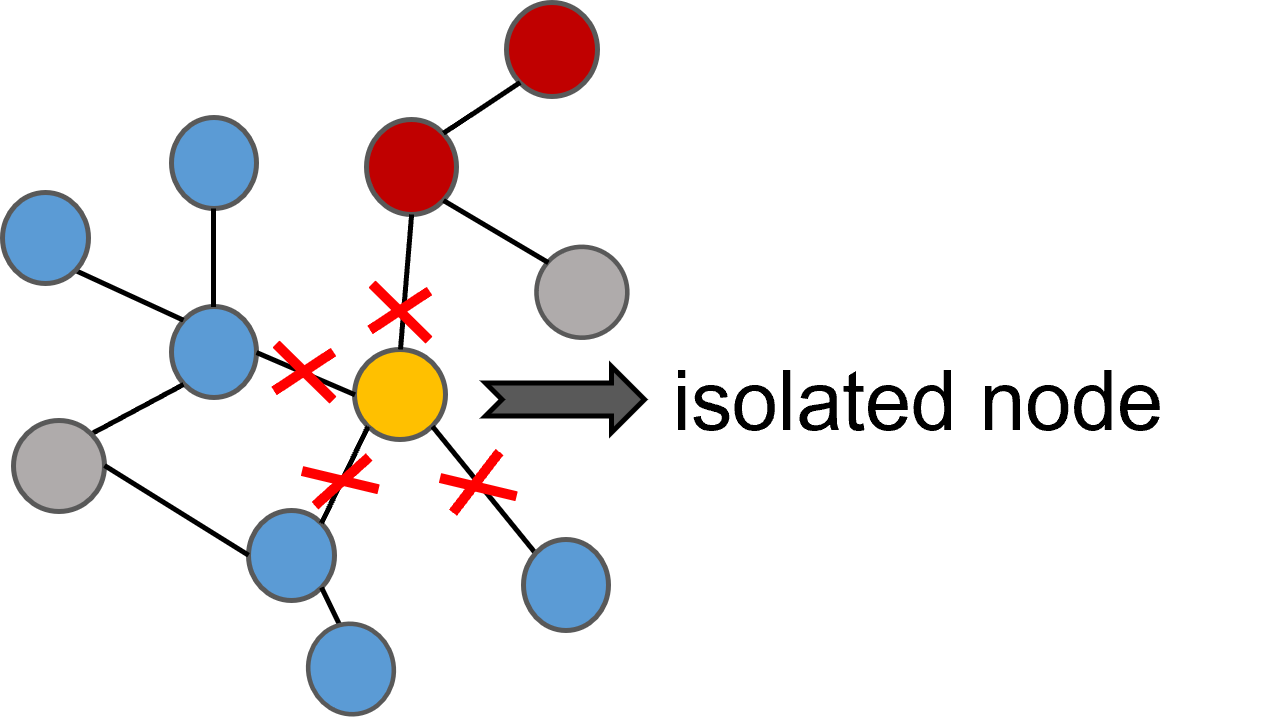}            
        \end{minipage}
        }
    \subfigure[Under-cutting]{
        \label{figb}
        \begin{minipage}[b]{0.4\linewidth}
            \centering
            \includegraphics[scale=0.4]{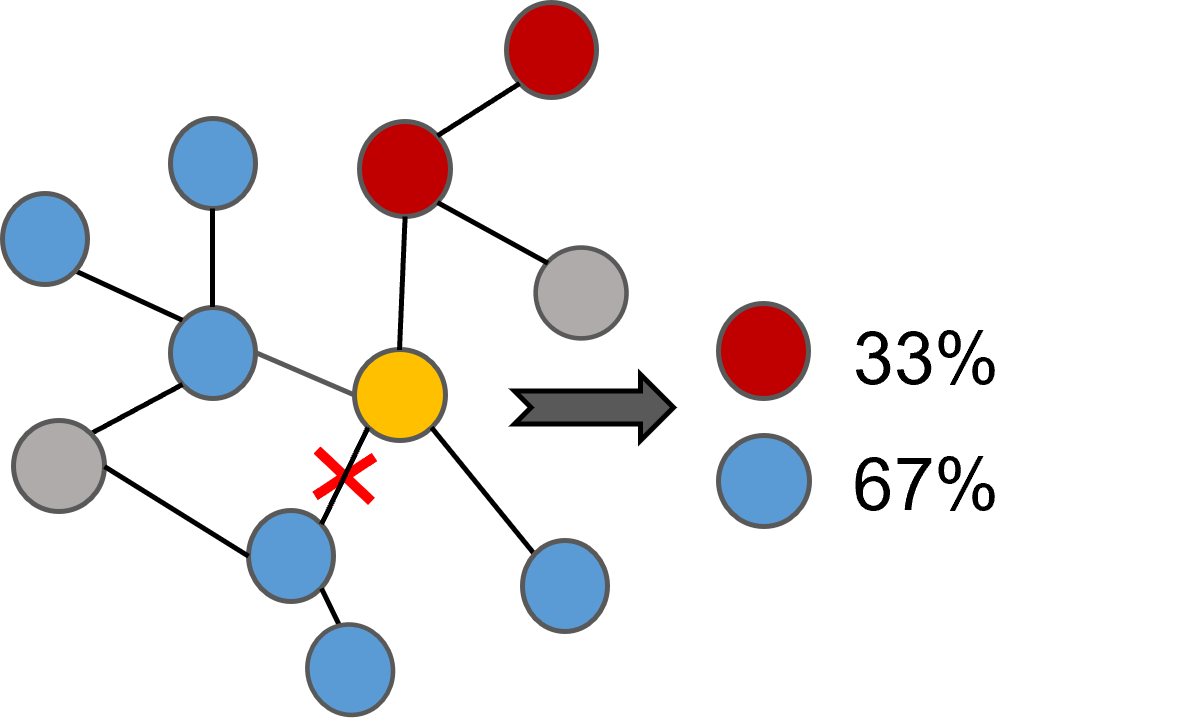}
        \end{minipage}            
        }
    \caption{Possible scenarios resulting from improper pruning edges of targets. }
    \label{fig:example}
\end{figure}

Cutting edges to address the influence of nodes with noisy labels in graph structures is effective, especially for outlier representation learning. However, the challenge is to decide the quantity of selected edges with high quality, as shown in Fig.~\ref{fig:example}. The examples claim that cutting excessively causes many isolated nodes, while too few edges fail to mitigate noisy information propagation. Faced with this difficulty, reinforcement learning~\cite{zhou_graphsr_2023,raman_learning_2021,wang_reinforced_2023} emerges as a promising approach due to efficient decision-making ability. Specifically, we utilize a tailored policy network to learn the strategy of selecting suspicious edges, taking node representations and graph structure represented by the adjacency matrix as the input. The output is the updated graph structure after refinement which needs evaluation from the detector. 

In this section, we reformulate the edge pruner module as a Markov Decision Process (MDP) to find the strategy for selecting edges based on the policy network $\pi_{\theta}$. The MDP is denoted as a tuple with $(\mathcal{S}, \mathcal{A}, \mathcal{P}, \mathcal{R}, \gamma)$, where $\mathcal{S}$ is the set of states of current graph structure, $\mathcal{A}$ is the set of actions of selecting candidate nodes and edges, $\mathcal{P}$ is the state transition function taking the policy network as a reference, $\mathcal{R}$ is the reward function, and $\gamma$ is the discounting factor. 

\noindent \textbf{Environment}: If this environment is not intervened manually, it would be the entire graph, a large-scale network of nodes and edges. However, it is evident that many nodes possess high credibility, making it unnecessary to take these nodes into consideration when searching for actions. Therefore, we design an initial, comparatively smaller environment. This setup is determined by a hyper-parameter, $\delta$, which does not play a decisive role, as the action selection process will further narrow the scope. We formulate nodes set in this initial environment as by $\mathcal{NC}$, i.e., 
\begin{equation}\label{eq:cand}
\mathcal{NC} = \left\{v_{i}|\hat{\mathrm{s}}_{i} \in \bar{\mathrm{s}}\pm \delta \right\}.
\end{equation}
where $\bar{\mathrm{s}}$ represents the average of anomaly scores for all nodes $V$. And $\hat{\mathrm{s}}_{i}$ is the predicted score for $v_{i}$. The detailed components are described as follows. 
\begin{itemize}
    \item \textbf{State} $\left(\mathcal{S}\right)$: The initial state $s_{0} = ( \hat{\mathrm{A}},\mathrm{H}^{L})$ comprises the adjacency matrix $\hat{\mathrm{A}}$ indicating the connectivity of the initial graph structure and node representations $\mathrm{H}^{L}$ from the $L$-layer base detector. In $t$-th step, state $s_{t-1}=(\hat{\mathrm{A}}_{t-1},\mathrm{H}_{t-1})$ is transitioned to the next state $s_{t} = \pi_{\theta}(a_t|s_{t-1})$ based on the edge-cutting action. 
    \item \textbf{Action} $(\mathcal{A})$: Selecting a target set including highly possible candidates with mistaken labels is significant. Candidates directly determine the search scope and quality of the pruner because incorrect labels occupy a small proportion of the graph. Thus, we define action $a_{t}=(\mathcal{N}_t, \mathcal{E}_t)$ includes two steps, choosing potential nodes $\mathcal{N}_t$ from $\mathcal{NC}$ by sampling and edges $\mathcal{E}_t$ centering the candidates, given current state $s_{t-1}$, as below:
    \begin{align}
    & \mathcal{N}_t = Sampling(\mathcal{NC})\\
    & \mathcal{E}_t =\left\{e_{ij}| i\in \mathcal{N}_{t}, j\in N_{i}, \sum_j n_{ij} \le n_e, \mathcal{N}_t \subseteq \mathcal{NC} \right\}, 
    \end{align}
    where $e_{ij}$ denotes the edges between $v_{i}$ and neighbors $N_{i}$. But the total number of $\sum_j n_{ij}$ is less than limitation $n_{e}$, as a hyper-parameter. The entire action search space is stated as follows:
    \begin{equation}
        \mathcal{A}_\mathcal{NC} = \left\{e_{ij}| v_{i}\in \mathcal{NC}, v_{j}\in N_{i}\right\} ,
    \end{equation}
    where $\mathcal{A}_\mathcal{NC}$ denotes all possible edge combinations of candidates in $\mathcal{NC}$. Besides, $\mathcal{E}_{t} \subseteq \mathcal{A}_\mathcal{NC}$ due to the edge limitation $n_{e}$, and $\mathcal{E}_{t}$ is closely related to the state transition probability function $\mathcal{P}$, explained in following Section~\ref{policynet}. Specifically, we utilize $\mathcal{E}_{T}= \sum_{t}^{T}\mathcal{E}_{t}$ to collect pruned edges from previous actions.  
    \item \textbf{Reward} $(\mathcal{R})$: Rewards play a vital role in guiding the policy network to choose actions efficiently and correctly. Specifically, we compute reward $r_{t}$ quantitatively indicated by comparing the metric of AUC (the area under the ROC curve) before and after $a_{t}$, that is:
\begin{equation}\label{reward}
 r_{t} = \mathcal{R}(s_{t-1}, s_{t}).
\end{equation} 
    The reward $r_{t}$ aims to evaluate the effectiveness after $a_{t}$. Moreover, the reward function $\mathcal{R}$ is formally and carefully designed by the improvement of the detector performance before and after taking $a_t$, under the AUC metric, relying on trustworthy pseudo labels $\mathcal{Y}^{'}$ instead of noisy ground truths as: 
\begin{equation}\label{eq:reward}
    r_{t}= \mathrm{AUC}(f_d(\mathcal{G}_{t}), \mathcal{Y}^{'}) - \mathrm{AUC}(f_{d}(\mathcal{G}_{t-1}),\mathcal{Y}^{'})
\end{equation}
where $\mathrm{AUC}(\cdot)$ is the function to compute performance and $f_{d}$ represents the base detector. Once the policy chooses $a_{t}=(\mathcal{N}_t,\mathcal{E}_t)$, including multiple edges centering candidate nodes in $\mathcal{N}_t$, the new graph $\mathcal{G}_{t} = (\mathcal{V},\mathcal{E} - \sum_t\mathcal{E}_t,X)$ can be easily obtained. 
\end{itemize}

The objective of the agent is to learn an optimal policy $\pi^{*}$ by targeting: 
\begin{equation}
    \arg\max_{\theta} \mathbb{E}_{\pi_{\theta}}\left[\sum_{t=1}^{T} \gamma^{t} r_{t} | s_{0}\right],
\end{equation}
where $s_{0}$ is the initial state. The process of edge pruning can be depicted as a trajectory $\left \{s_{0},a_{1},s_{1},r_{1},a_{2},s_{2},...,a_{T},s_{T}, r_{T}\right\}$, consisting of $T$ steps. This is the complete definition of the Markov decision process, and next, we explain how to apply the base detector to compute rewards given the graph structures before and after refinement in detail.

\subsection{Base Detector} \label{detector}  
The base detector $f_{d}$ is proposed to predict anomaly scores and generate pseudo labels, especially for unlabeled data. Although this base detector is supervised by noisy ground truth labels $\bar{\mathcal{Y}}_{L}$, the detector, which generates pseudo labels for unlabeled nodes and provides guidance for reward design dependent on confident predictions, plays a vital role in the \textit{policy-in-the-loop} mechanism. Due to not all predictions being reliable, we identify two trustworthy sets: anomaly node set $\mathcal{AS}$ and normal node set $\mathcal{NS}$ and consider confident predictions as their reliable truths. 

The base detector employs the loss function from Meta-GDN model~\cite{gao_alleviating_2023} not only because this model provides superior performance in GAD task but also addresses data imbalance in training process~\cite{gao_addressing_2023} through balanced batches. $f_d$ maps node features into the low-dimensional latent space to learn node representations. We apply the simple Graph Convolutional Network (GCN) to learn node embeddings for score evaluation. To represent aggregation formally, the computation with an activation function of layer $l$ is written as: 
\begin{equation}\label{eq:gcn}
\mathrm{H}^{l} = \sigma \left ( \hat{\mathrm{A}}\mathrm{H}^{l-1} \mathrm{W}^{l-1}\right ),
\end{equation}
where $\hat{\mathrm{A}} = \tilde{\mathrm{D}}^{-1/2} \tilde {\mathrm{A}}\tilde{\mathrm{D}}^{-1/2}$ and  $\tilde {\mathrm{A}} = \mathrm{A}+\mathrm{I}$. To make it clear, $\mathrm{I}$ and $\mathrm{D}$ are the identity matrix and the degree matrix separately. $\sigma (\cdot)$ denotes the activation function, and the simplified expression is shown in Eq.~\ref{eq:simplegcn}. Graph neural networks are usually designed with multiple layers $L$ to keep the long-range information, and thus node representations are $\mathrm{H}^{L}\in \mathbb{R}^{d \times r}$.
\begin{equation}\label{eq:simplegcn}
\mathrm{H}^{l} =  GCN(\hat{\mathrm{A}},\mathrm{H}^{l-1}).
\end{equation}

Given embeddings, the detector evaluates an anomaly score for each node, namely the probability of being anomalous nodes. The score evaluation is computed by a simple linear layer based on $\mathrm{H}^{L}$ as follows:
\begin{equation}\label{eq:liner}
\hat{\mathrm{S}} = \mathrm{W}\mathrm{H}^{L} + b, 
\end{equation}
where $\hat{\mathrm{S}}$ denotes the predicted scores. And $\hat{\mathrm{s}}_{i}$ denotes the score of node $v_{i}$. If $\hat{\mathrm{s}}_{i}\approx 1 $ $v_{i}$ is anomalous, vice versa. However, the score $v_{i}$ is difficult to discern when $\hat{\mathrm{s}}_{i}\approx \bar{\mathrm{s}}$, which ensembles a potential candidate with false labels. 


By ranking $\hat{\mathcal{S}}$, two high-confidence sets are filtered out based on a hyper-parameter rate $\alpha$, i.e., anomaly set $ \mathcal{AS}=\left \{ v_{1}, ...v_{N_{as}} \right \}$ (the most suspicious anomalies), normal set $\mathcal{NS}=\left \{ v_{1}, ...v_{N_{ns}} \right \}$ (the most reliable normal nodes). They are selected for label rectification to guarantee fewer incorrect ground-truth labels and better supervision for training, i.e.,
\begin{align}\label{eq:thres}
    \mathcal{AS}&= \left\{v_{i}|\hat{\mathrm{s}}_{i} \in f_{top}^{\alpha}(\hat{\mathrm{S}})\right\}, \\
    \mathcal{NS}&= \left\{v_{i}|\hat{\mathrm{s}}_{i} \in f_{top}^{\alpha}(-\hat{\mathrm{S}})\right\},
\end{align}
where $f_{top}$ is the filtering function by ranking scores. Next, the predicted labels of $\mathcal{AS}$ and $\mathcal{NS}$ are treated as trustworthy ground truth labels, i.e.,
\begin{equation}\label{eq:revise}
\mathcal{Y}^{'} =
\begin{cases}
  1 & \text{ if } v_{i} \in \mathcal{AS}; \\
  0 & \text{ else if } v_{i} \in \mathcal{NS};\\
  \bar{y}_i &\text{ else if } v_i \in \mathcal{V}_{L}; \\
  y_{i}^{\prime} &\text{else,}
\end{cases}
\end{equation}
where $\mathcal{Y}^{'}$ denotes more reliable ground-truth labels than the initial $\bar{\mathcal{Y}}_{L}$. Besides, the base detector is pre-trained during the whole optimization.

\subsection{Policy-in-the-loop}\label{loop}
The \textit{policy-in-the-loop} frame is depicted as a cyclic framework where the base detector and pruner interact and share feedback mutually to finish the task of removal of suspicious edges. This loop plays a role in controlling the policy to determine the edge number to prune, as well as ensuring the quality of pruned edges to achieve the goal of controlling noisy information propagation. Concretely, the base detector assigns anomaly scores by learning from the refined graph providing a reasonable basis for reward calculation. In addition, the edge pruner takes actions to address noisy-label information transmission providing graph structure for the detector in each iteration. 

\subsubsection{Policy Network}\label{policynet}
We describe the process of the policy network to estimate the edge probability to prune determining action $a_{t}$ specifically. The policy network leverages $L$ GCN layers, as below:
\begin{equation}
    \begin{aligned}
    \mathrm{P}_{t} & = GCN^{L}(\hat{\mathrm{A}}_{t},\mathrm{H}_{t}),
    \end{aligned}
\end{equation}
where $GCN(\cdot)$ represents the simplized formula as Eq.~\ref{eq:simplegcn}. The output $\mathrm{P}_{t}$ is a probability matrix, where $p_{ij}$ represents the likelihood of $e_{ij}$ to be removed. 

\subsubsection{State Transition}
The action $a_{t}$ selects multiple edges simultaneously by selecting the top-rank samples based on $\mathrm{P}_{t}$ within limitation $n_{e}$, and thus the graph structure is updated, i.e., 
\begin{equation}\label{selection}
    \pi_{\theta}(a_t|s_{t-1}) = (Bernoulli(v_i), \bigcup_{v_i \in \mathcal{N}_t} \arg\max_{e_{ij}}( p_{ij}, j\le n_e) )
\end{equation}
and the state transition probability from $s_{t-1}$ to $s_{t}$ is written as: 
\begin{equation}\label{transition}
p(s_{t}) = \pi_{\theta}(a_t|s_{t-1})(\mathrm{P}_{t} \odot \mathrm{M}_{t}),
\end{equation}
where $\mathrm{M}_t$ represents a mask matrix to exclude those already pruned edges$\mathcal{E}_{t-1}$ and edges not connected with nodes in $\mathcal{N}_t$. 
The function $\mathcal{P}$ involves two steps: i) $\hat{\mathrm{A}}_{t-1}$ is modified corresponding to the selected edges $\mathcal{E}_t$, resulting in a new adjacency matrix $\hat{\mathrm{A}}_{t}$; ii) The node embeddings $\mathrm{H}_{t}$ are obtained by passing the modified $\hat{\mathrm{A}}_{t}$ to the base detector. 

\subsubsection{Policy Gradient Learning}
To optimize the policy network, the objective function is formulated to maximize the cumulative rewards during the MDP with the REINFORCE method, i.e., 
\begin{equation}\label{policy_loss}
     \mathcal{J}_{\theta} = - \sum_{t}^{T} \sum_{e_{ij}}^{\mathcal{E}_{t}}p_{ij} \pi_ {\theta}(a_t|s_{t-1}) r_{t} \gamma_{t},
\end{equation}
where $\mathcal{J}_{\theta}$ denotes the objective function during training the policy network and $p_{ij}$ is the cutting probability values of edge $e_{ij}$. $\gamma_{t}$ is the discount factor, and $r_{t}$ is the base detector performance after executing a batch cut $\mathcal{E}_{t}$ at the $t$-th step. 

The quantity of pruned edges is crucial as over-cutting may lead to numerous isolated nodes, while under-cutting may fail to reduce noisy information propagation. Therefore, the terminal condition is set to ensure balanced pruning as:
\begin{equation}\label{eq:terminal}
    N_{\mathcal{E}_{T}} \le N_{nc}~ and ~\mathcal{E}_{t} \to Null,
\end{equation}
where $N_{\mathcal{E}_{T}} $ represents the edge sum in all previous actions, and $N_{nc}$ is the edge count surrounding nodes in the set $\mathcal{NC}$. This ensures the policy network finds a balance in edge pruning to reduce the negative impacts of nodes with noisy labels. 

\section{Experiments}

In this section, we conduct experiments to evaluate the performance of REGAD and answer the following questions: 

\noindent \textbf{RQ1:} How effective is the proposed model in graph anomaly detection under the noisy-label setting compared to baselines?

\noindent \textbf{RQ2:} How does the REGAD perform under different noisy label ratios and anomaly node ratios across three datasets? 

\noindent \textbf{RQ3:} How do policy backbones affect model performance?  

\noindent\textbf{RQ4:} Hyperparameters analysis, including the confident pseudo-label ratio $\alpha$ and the edge limitation $n_e$?


\begin{table*}[htbp] 
  \centering
  \renewcommand\arraystretch{1.25}
  \caption{Performance of REGAD and baselines under the same noisy label ratio of 50\% (among labeled anomalies).}
    \begin{tabular}{ccccccc}
    \toprule
    \multirow{2}{*}{\textbf{Methods}} & \multicolumn{2}{c}{\textbf{Clothing}} & \multicolumn{2}{c}{\textbf{Computer}} & \multicolumn{2}{c}{\textbf{Photo}} \\
    \cmidrule(r){2-3} \cmidrule(r){4-5} \cmidrule(r){6-7}
   & \textbf{AUC} & \multicolumn{1}{c}{\textbf{AUPR}} & \multicolumn{1}{c}{\textbf{AUC}} & \multicolumn{1}{c}{\textbf{AUPR}} & \multicolumn{1}{c}{\textbf{AUC}} & \multicolumn{1}{c}{\textbf{AUPR}} \\
    \midrule
    DOM   & 0.502±0.001 & 0.015±0.001 & 0.500±0.002 & 0.020±0.001 & 0.494±0.005 & 0.021±0.002\\
    ComGA & 0.537±0.015 & 0.041±0.002 & 0.539±0.017 & 0.047±0.002 & 0.481±0.026 & 0.060±0.060\\
    \hline
    BWGNN  & 0.512±0.003 & 0.044±0.003 & 0.501±0.002 & 0.044±0.001 & 0.501±0.002 & 0.045±0.001 \\
    GHRN  & 0.513±0.001 & 0.046±0.003 & 0.499±0.001 & 0.043±0.001 & 0.502±0.003 & 0.045±0.001 \\
    DeepSAD & 0.559±0.013 & \textbf{0.150±0.009} & 0.533±0.057 & 0.066±0.043 & \underline{0.914±0.008}& \underline{0.452±0.045} \\
    Meta-GDN & \underline{0.614±0.047} & 0.094±0.010 & 0.705±0.095 & 0.261±0.024 & 0.876±0.020 & 0.421±0.007 \\
    \hline
    RTGNN & 0.501±0.015 & 0.034±0.002 & 0.397±0.018 & 0.033±0.001 & 0.524±0.018 & 0.046±0.002 \\
    D2PT  & 0.537±0.002 & 0.048±0.001 & \underline{0.751±0.008} & \underline{0.328±0.018} & 0.786±0.057 & 0.186±0.081 \\
    PIGNN & 0.414±0.014 & 0.026±0.001 & 0.470±0.045 & 0.039±0.005 & 0.512±0.022 & 0.046±0.002 \\
    \hline
    REGAD & \textbf{0.685±0.006} & \underline{0.110±0.003} & \textbf{0.767±0.005} & \textbf{0.357±0.014} & \textbf{0.924±0.006} & \textbf{0.486±0.027} \\ 
    \bottomrule
    \end{tabular}%
  \label{tab:comprehensive}%
\end{table*}%

\subsection{Experimental Settings}
\noindent\textbf{Datasets.}
We adopt three real-world attributed graphs that have been widely used in related studies, i.e., Clothing~\cite{ding_graph_2020}, Computer~\cite{shchur_pitfalls_2019}, and Photo~\cite{shchur_pitfalls_2019}. We follow the standard setup of graph anomaly detection research~\cite{zhou_Improving_2023} and consider nodes from the smallest class(es) as anomaly data (i.e., rare categories) while nodes from the other classes as "normal" data. 

The details of three real-world datasets are as follows: 
\begin{itemize}
\item \textbf{Clothing}~\cite{ding_graph_2020}: This dataset includes items such as clothing and jewelry from the Amazon website as nodes. The edges in this network represent instances in which two products are purchased together.  
\item \textbf{Computer}~\cite{shchur_pitfalls_2019}: This dataset is a segment of the co-purchase graphs about computer-related items. Nodes represent individual device products, and edges represent frequent purchasing behaviors.  
\item \textbf{Photo}~\cite{shchur_pitfalls_2019}: This dataset focuses on photography-related products, similar to the Computer dataset. 
\end{itemize} 

\begin{table}[htbp]
\centering
    \caption{Statistics of datasets and $r$ denotes the anomaly ratio.}
    \setlength{\tabcolsep}{3.5mm}
    \begin{tabular}{cccccc}
    \toprule
    \textbf{Datasets}  & \textbf{Clothing} & \textbf{Computer} & \textbf{Photo}   \\
    \midrule
        \# Features & 9,034 & 767   & 745     \\
        \#  Nodes & 24,919  & 13,381 & 7,487  \\
        \# Edges  & 91,680 & 245,778 & 119,043 \\
        \# Anomalies & 856 & 580   & 331      \\
        \textbf{$r$}  & 3.44\% & 4.33\% & 4.42\%  \\
    \bottomrule
    \end{tabular}
\label{tab:dataset}
\setlength{\belowdisplayskip}{-0.5pt}
\vspace{-0cm}
\end{table}

\noindent\textbf{Baselines.}
We compare REGAD with three-group methods for effectiveness evaluation. The three types include: (1) unsupervised GAD methods (e.g., DOM~\cite{ding_deep_2019} and ComGA~\cite{luo_ComGA_2022}), (2) semi-supervised GAD methods (e.g., BWGNN~\cite{tang_rethinking_2022}, CHRN~\cite{gao_addressing_2023}, DeepSAD~\cite{ruff_deep_2020}, Meta-GDN~\cite{ding_Fewshot_2021}, ), (3) noisy label learning methods (e.g., RTGNN~\cite{qian_Robust_2023}, D2PT~\cite{liu_learning_2023}, PIGNN~\cite{du_Noiserobust_2021}). 

The details of the baselines are as follows:
\begin{itemize}
    \item DOM~\cite{ding_deep_2019} is an unsupervised method that includes a decoder and an encoder by reconstructing the adjacency matrix to detect anomalies. 
    \item ComGA~\cite{luo_ComGA_2022} designs a community-aware method to obtain representations to predict anomaly scores. 
    \item BWGNN~\cite{tang_rethinking_2022} is proposed to address the `right-shift' phenomenon of graph spectrum by Beta Wavelet Graph Neural Network. 
    \item CHRN~\cite{gao_addressing_2023} addresses the heterophily of the GAD task by emphasizing high-frequency components by the graph Laplacian. 
    \item DeepSAD~\cite{ruff_deep_2020} evaluates the entropy of the latent distribution for normal nodes and anomalous nodes to classify.
    \item Meta-GDN~\cite{ding_Fewshot_2021} is the simple version of an anomaly detection model based on GCN, leveraging balanced batch sizes and deviation loss between abnormal and normal nodes.  
    \item RTGNN~\cite{qian_Robust_2023} focuses on noise governance by a self-reinforcement supervision module and consistency regularization after graph augmentation.  
    \item D2PT~\cite{liu_learning_2023} innovates a dual-channel GNN framework, increasing robustness considering the augmented and original global graphs. 
    \item PIGNN~\cite{du_Noiserobust_2021} leverages structural pairwise interactions (PI) to propose a PI-aware model to manage noise.
\end{itemize}

\noindent \textbf{Evaluation metrics}\label{evaluation}
Following widely used evaluation metrics~\cite{ding_Fewshot_2021,zhou2022unseen}, we adopt two metrics, i.e., AUC and AUPR, which have been widely used. AUC denotes the area under the ROC curve, which illustrates the true positive rate against the false positive rate. AUPR is the area under the Precision-Recall curve, showing the trade-off between precision and recall. 

\noindent \textbf{Implementation Details} \label{implement}

All datasets are split into training (40\%), validation (20\%), and test (40\%) sets. Our research problem includes noisy labels, which are difficult to distinguish during the training and validation phases. However, current datasets are pre-processed and basically contain clean labels. To simulate the real-world scenarios of noisy labels, we follow related papers~\cite{zhao_ADMoE_2022,dai_NRGNN_2021}. Therefore, we induce noisy labels into the datasets by label flipping and then mixing corrupted labels with correct ones. Erroneous labels for outliers are generated by uniformly swapping labels of normal nodes as anomalous nodes at a designated rate. 

In the experiments, if not further specified, the number of labeled anomalies is set as 5\%, and the noisy label ratio is 50\% of all labeled anomalous nodes. Moreover, we adopt a pre-trained base detector with 2 layers of GCN with 128 hidden units to learn node representations and one linear layer to compute anomaly scores. Similarly, we leverage a 2-layer GCN as the policy network to manipulate edge-cutting probabilities for edges. The learning rate of the policy network is 0.005, and the weight decay is set to 5e-4. 
To make $\mathcal{AS}$  and $\mathcal{NS}$ more reliable, the rate $\alpha$ is a significant hyperparameter ranging in [0.001, 0.01]. We filter out edges for nodes in $\mathcal{N}_t$ simultaneously. Edge sampling limitation $n_{e}$ for candidates is set in [100, 150]. Episode $T$ ranges \{5,10,15,20\} to optimize the policy network. We train REGAD with 5-10 epochs to ensure stability in one seed. We run ten experiment seeds and report the average results. 

\subsection{Effectiveness Analysis (RQ1)}
We present the results of AUC and AUPR in Table~\ref{tab:comprehensive} and have the following observations. Semi-supervised models (e.g., DeepSAD) marginally outperform the unsupervised models (e.g., ComGA). This suggests that supervised ground truth, even with noisy labels, can still guide learning anomaly and normal node patterns. So in our model, we keep the noisy ground truth as weak supervision and attempt to rectify some mistaken truths with reliable pseudo-labels. The noisy label learning methods (e.g., RTGNN and D2PT) based on graph neural networks (GNNs) achieve sub-optimal performances on these three datasets compared with the semi-supervised models. This verifies that the existing GNN-based methods exploited to improve robustness under noisy labels are unsuitable for the imbalanced GAD task with noisy labels.

The performance of our method, REGAD, significantly surpasses baselines in a relatively highly noisy label scenario. Inspired by semi-supervised and noisy label learning methods, REGAD relies less on true labels and mitigates the influence of noisy labels from the graph topology. This is because REGAD refines the graph structure to control noise propagation and utilize confident predictions as ground truth simultaneously, which helps to provide more reliable supervision for learning outlier patterns in the \textit{policy-in-the-loop} framework. 

\subsection{Robustness Analysis(RQ2)}\label{noise}

\begin{figure}[htbp]
    \centering       
    \includegraphics[width=1\linewidth,trim = 0.45cm 0.7cm 0.3cm 0.2cm,clip]{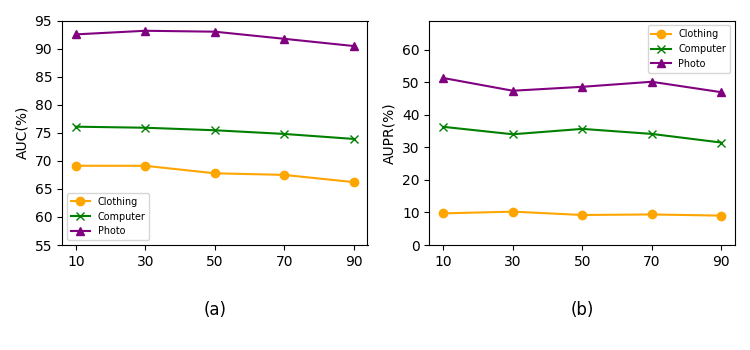}
    \caption{Impact analysis of different noisy label ratios among the same labeled anomalous nodes estimated by AUC and AUPR.}
    \label{fig:noisy}
\end{figure}

In this section, we compare the model performance by implementing experiments with different noisy label ratios, i.e., \{10\%, 30\%, 50\%, 70\%, 90\%\} among labeled anomalies to analyze the robustness and effectiveness of REGAD. Firstly, we investigate the impacts of different noisy label ratios under 5\% fixed labeled anomalous nodes. In Fig.~\ref{fig:noisy}, the results based on the AUPR metric are consistent despite scale variations of different datasets. The general trend of REGAD performance is downward as the noisy label ratio increases, which aligns with noisy-label impacts. However, fluctuations are observed: for instance, there is an increase followed by a continuous decrease from 30\% to 50\%, evident in the Computer and Photo dataset. Introducing noisy labels within a proper range (30\%-50\%) may have a regularization effect, enabling REGAD to learn anomaly patterns and make accurate decisions to prune edges. Additionally, because the AUPR metric is more sensitive to predictions of minority classes, our model exhibits greater robustness at the moderate noisy label ratios. The Clothing dataset remains stable generally, which demonstrates REGAD's effectiveness on sparse graphs. 

\begin{figure}[htbp]
    \centering       
    \includegraphics[width=1\linewidth,trim = 0.45cm 0.7cm 0.3cm 0.2cm,clip]{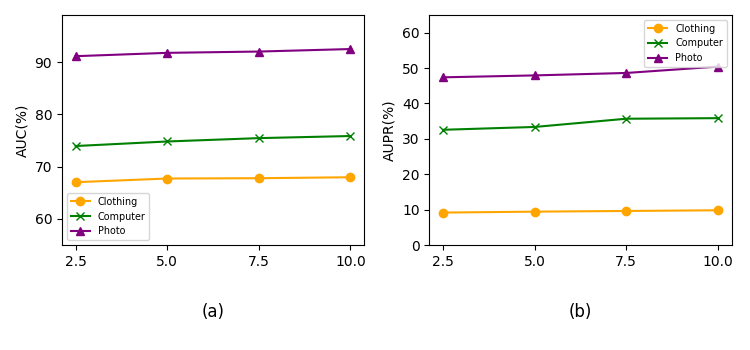}
    \caption{Impact analysis of the labeled anomalous node ratios with the same noisy label percentage estimated by AUC and AUPR.}
    \label{fig:label}
\end{figure}

Under different known labeled outliers, we set labeled ratios as \{2.5\%, 5\%, 7.5\%, 10\%\} respectively with the same ratio (50\%) among labeled nodes. Under the AUC in Fig.~\ref{fig:label}, when labeled anomaly nodes increase, the performance of the three datasets improves because the weak supervision from noisy ground truths still functions well. 
In extreme cases, such as with only 2.5\% labeled anomalous nodes, REGAD achieves relatively good results. This validates that the design of \textit{policy-in-the-loop} framework, including treating confident predictions as truths by the detector and cutting edges to control noise propagation, is superior. The model exhibits a similar trend when evaluated using AUPR. 
\subsection{Ablation Study (RQ3)}
Due to our design of a policy-in-the-loop framework, we aim to further investigate how policies based on different bases affect the overall performance. Therefore, we designed the policy network using four different structures: MLP, GCN, GAT, and GraphSAGE, to examine the influence of various architectures on the policy's effectiveness.

According to the results (see Table~\ref{tab:policy_results}), the policy based on GAT demonstrates the best performance consistently across all three datasets. In contrast, MLP shows the weakest effect, likely due to its limited learning capacity and lack of specialization for edge selection between different nodes.

\begin{table}[htbp]
\centering
\caption{Performance impacts of different layers used in policy networks of the pruner.}
\label{tab:policy_results}
\begin{tabular}{lccc}
\toprule
Policy & Clothing & Computer & Photo \\
\midrule
No-cutting & 0.633  & 0.727  & 0.892 \\
MLP & 0.672  & 0.743  & 0.895 \\
GAT & 0.685 & 0.765 & 0.923 \\
GraphSAGE & 0.682 & 0.766 & 0.923 \\ 
GCN & \textbf{0.685} & \textbf{0.767} & \textbf{0.924} \\ 
\bottomrule
\end{tabular}
\end{table}

\subsection{Sensitivity Analysis (RQ4)}

\begin{figure*}[htbp]
    \small 
    \centering
    \subfigure[$\alpha$ on Clothing]{
        \label{fig:1}
        \begin{minipage}{0.3\linewidth}
            \centering
            \includegraphics[width=1\linewidth]{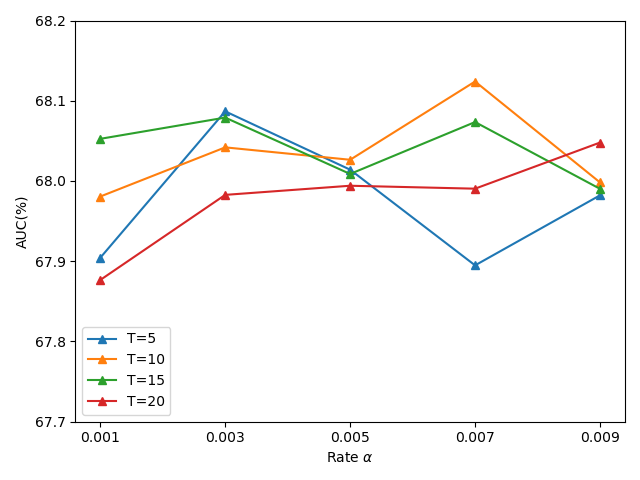}        
        \end{minipage} 
    }
    \subfigure[$\alpha$ on Computer]{
        \label{fig:2}
        \begin{minipage}{0.3\linewidth}
            \centering
            \includegraphics[width=1\linewidth]{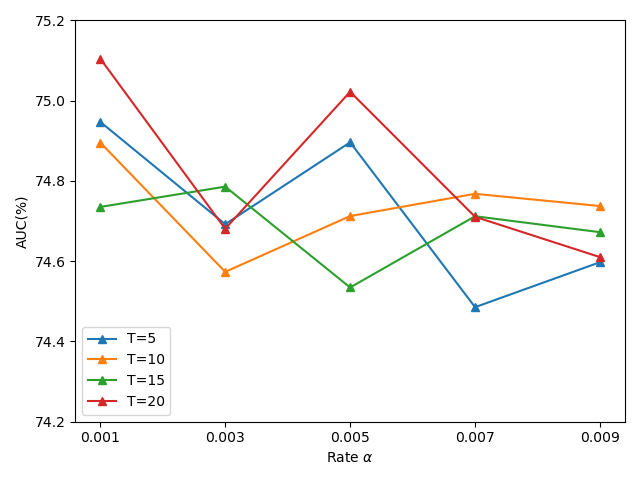}
        \end{minipage}
    } 
    \subfigure[$\alpha$ on Photo]{
    \label{fig:3}
        \begin{minipage}{0.3\linewidth}
        \centering
        \includegraphics[width=1\linewidth]{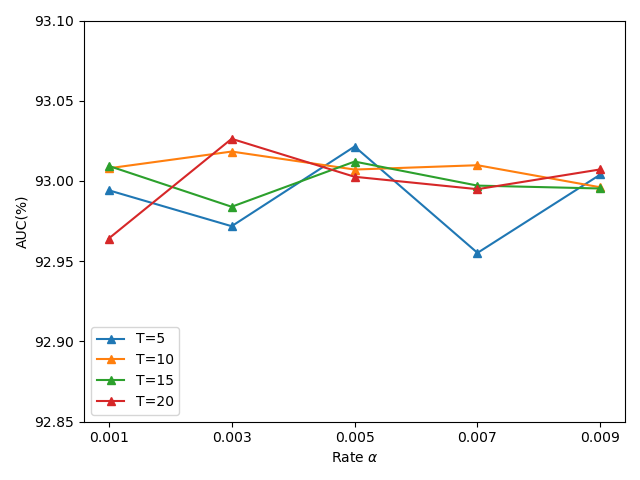}
    \end{minipage}
    }

    \subfigure[$n_e$ on Clothing]{
    \label{fig:4}
        \begin{minipage}{0.3\linewidth}
            \centering
            \includegraphics[width=1\linewidth]{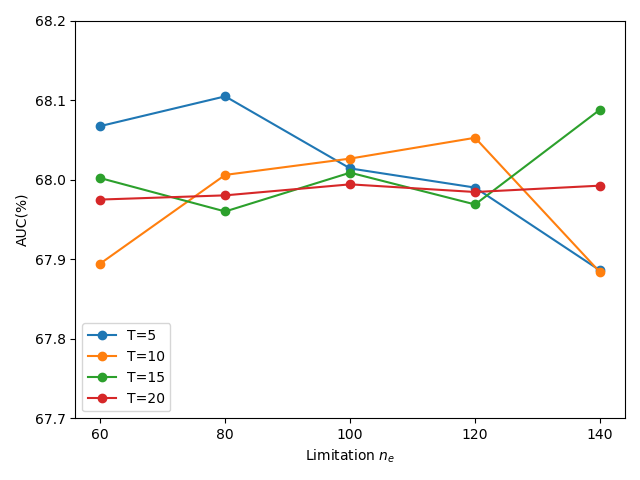}
    \end{minipage}
    } 
    \subfigure[$n_e$ on Computer]{
    \label{fig:5}
    \begin{minipage}{0.3\linewidth}
        \centering
        \includegraphics[width=1\linewidth]{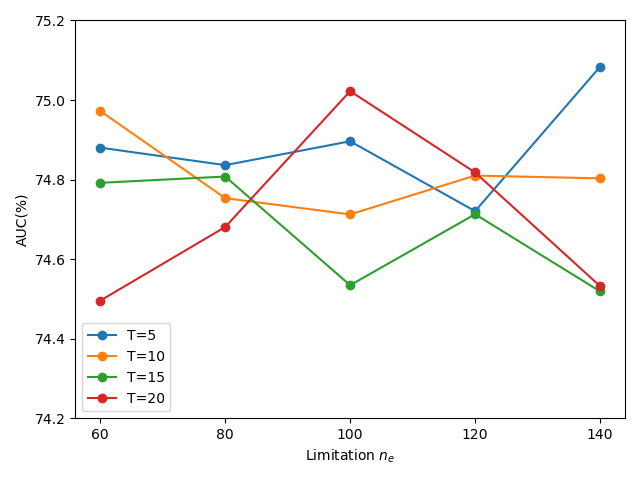}
    \end{minipage}
    }
    \subfigure[$n_e$ on Photo]{
    \label{fig:6}
    \begin{minipage}{0.3\linewidth}
        \centering
        \includegraphics[width=1\linewidth]{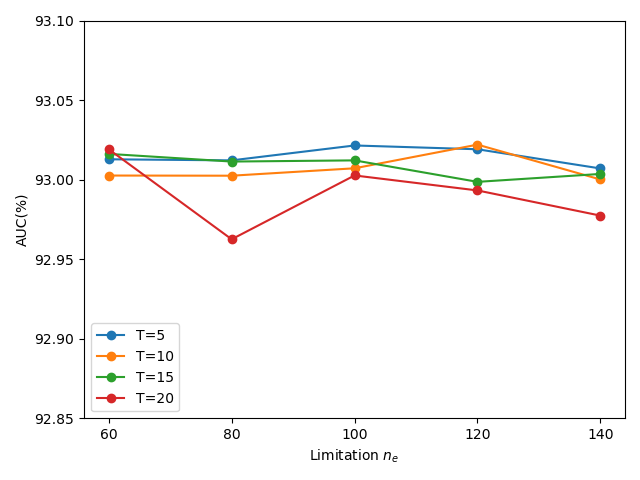}
    \end{minipage}
    }
\caption{Sensitivity analysis of label refined ratio $\alpha$ and edge limitation $n_e$ on three datasets.}
\label{fig:hyper}
\end{figure*}

This section establishes whether our model is sensitive to two key hyperparameters: i) $\alpha$, determining how many pseudo predictions are employed as confident truths rather than using the initial noisy labels, and ii) $n_{t}$, the constraint of selected edges in each action. The performance of hyperparameters on the Clothing dataset is shown in Fig.~\ref{fig:hyper}. Furthermore, the value ranges of the two parameters are \{0.001, 0.003, 0.005, 0.007, 0.009\}, \{60,80,100,120,140\} respectively. 
Specifically, in Fig.~\ref{fig:1} and Fig.~\ref{fig:4}, we can observe that REGAD is not very sensitive to this $\alpha$ on the Clothing dataset. For the $\alpha$, smaller values (e.g., 0.001-0.005) are generally recommended because $\alpha$ determines the proportion of labels directly modified based on the estimation of the base detector. If this proportion is too high, it might introduce noise, requiring more actions to correct newly generated noisy labels. $n_{e}$ determines the maximum allowable number of edges to be selected in action, enabling the gradual and regulated steps for edge selection. The value is recommended in the middle level (i.e., 80-100). Conversely, the limitation is widened to 140, and the performance falls sharply in most scenarios. This suggests that our model is stable and robust, reaffirming its effectiveness across varied settings. In conclusion, the findings highlight the reliability of REGAD in graph anomaly detection tasks. 

Fig.~\ref{fig:2} and Fig.~\ref{fig:5} display that the Computer shows higher sensitivity to the hyperparameter values. The recommended settings suggest $\alpha$ as 0.001 and $n_e$ between 100 and 140. However, for the Photo dataset in Fig.~\ref{fig:3} and Fig.~\ref{fig:6}, the sensitivity is relatively low, with minimal performance fluctuations. It is recommended to set alpha between 0.003 and 0.005, and $n_e$ is generally suggested to be set between 100 and 120, which would be appropriate. 

\subsection{Case Study And Efficiency Analysis}

\begin{figure*}[htbp]
    \centering
    \subfigure[Clothing]{
    \label{fig:case_clothing}
    \begin{minipage}{0.31\linewidth}
         \centering
        \includegraphics[width=1\linewidth]{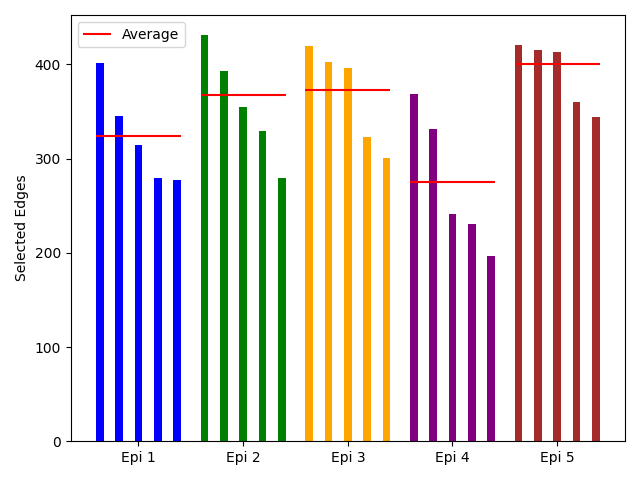}
    \end{minipage}}
    \hfill
    \subfigure[Computer]{
    \label{fig:case_computer}   
    \begin{minipage}{0.31\linewidth}
         \centering
        \includegraphics[width=1\linewidth]{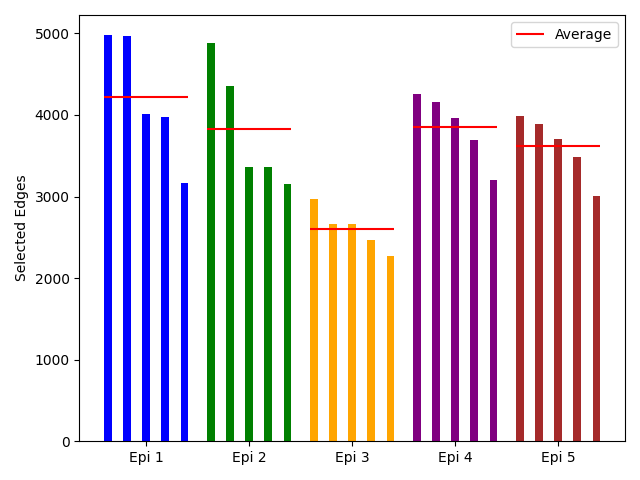}
    \end{minipage}
    }
    \hfill
    \subfigure[Photo]{
    \label{fig:case_photo}
    \begin{minipage}{0.31\linewidth}
         \centering
        \includegraphics[width=1\linewidth]{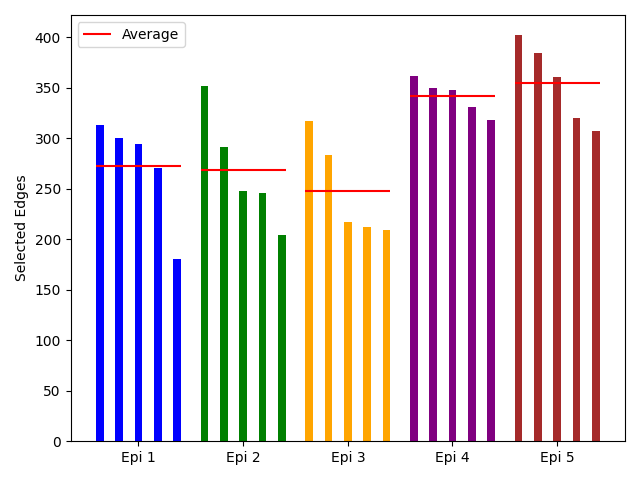}
    \end{minipage}
    }
\caption{The number of selected edges of the first five actions for 5 episodes on three datasets.}
\label{fig:casestudy}
\end{figure*}

To further investigate how the edge pruner functions effectively, we analyze selected edges in actions when different episodes and epochs are set. This case study explores how the number of edges selected in each MDP in Fig~\ref{fig:casestudy}. We aim to understand how the pruner completes the graph refinement process in several steps and whether the agent exhibits different strategies across episodes. We observe some interesting phenomena. In short, the listed examples reveal that REGAD has an efficient, balanced edge pruning ability to mitigate the influence of noisy labels for anomaly pattern learning. 

The pruned edge average is slightly different among different episodes but indicates a stable performance among actions within one episode. The variations are potentially attributed to the process of policy network optimization, where the model continuously explores different edge-pruning strategies. Conversely, actions with one episode tend to be reduced step by step. The possible reason could be the model's adaptation to newly learned patterns, and the available candidate edges are pruned.

To further demonstrate how the edge pruner modifies the overall label relationship by adjusting the graph structure, we present the number of edges pruned and labels modified across three datasets in Table~\ref{tab:edge_pruning}. Specifically, the pruned edges reflect the removal of potentially less informative or redundant connections between nodes, which contributes to a more compact and relevant graph representation. Meanwhile, the modified labels indicate how the pruning process indirectly influences the label propagation or correction, showcasing the pruner's impact on downstream predictions.

By comparing these numbers across datasets, we observe dataset-specific pruning behaviors — for instance, the Photo dataset exhibits the highest number of pruned edges but comparatively fewer label modifications, suggesting a more aggressive edge filtering with limited label adjustment. In contrast, the Clothing dataset shows a balanced count of pruned edges and label modifications, reflecting a trade-off between structural simplification and label refinement. 
\begin{table}[ht]  
\centering  
\caption{Statistics of removed edges for each action on average and refined ground-truth labels before cutting. } 
\begin{tabular}{lcc}  
\hline  
\textbf{Dataset} & \textbf{Pruned Edges} & \textbf{Modified Labels} \\
\hline  
Clothing & 228 & 28 \\
Computer & 3692 & 36 \\
Photo & 380 & 17 \\
\hline  
\end{tabular}  
\label{tab:edge_pruning}  
\end{table}

We evaluated the computational time on three datasets. Owing to the integration of GNN and reinforcement learning, our model demonstrates significantly improved computational efficiency.  
The results in Table~\ref{tab:dataset_timings} show that the Clothing dataset requires substantially more training and testing time compared to the Photo and Computer datasets. This difference likely stems from the varying dataset sizes and complexities. 

\begin{table}[htbp]  
\centering  
\caption{Dataset Training, Validation, and Testing Time (seconds)}  
\begin{tabular}{lccc}  
\hline  
\textbf{Dataset} & \textbf{Train (s)} & \textbf{Valid (s)} & \textbf{Test (s)} \\
\hline  
Clothing & 417.46 & 0.11 & 56.73 \\
Photo & 16.44 & 0.02 & 1.62 \\
Computer & 6.75 & 0.08 & 1.05 \\
\hline  
\end{tabular}  
\label{tab:dataset_timings}  
\end{table}
\section{Related work}
\subsection{Graph Anomaly Detection}
Graph anomaly detection (GAD) is a crucial task to identify graph objects that deviate from the main distribution~\cite{jiang_weakly_2023} with inherent connections and complex structures. Previous GAD methods roughly fall into GNN-based and deep-learning methods. 
GNN-based methods~\cite{ding_interactive_2019} utilize graph topology and manage neighborhood information, i.e., local node affinity~\cite{qiao_Truncated_2023}. A benchmark~\cite{tang_gadbench_2023} illustrates that simple supervised ensembles with neighborhood aggregation also perform well on GAD. Unsupervised graph auto-encoder methods~\cite{ding_deep_2019,fan_anomalydae_2020} encode graphs and reconstruct graph structure by the decoder to detect anomalies. Some research leverages anomalous subgraphs~\cite{liu_SelfInterpretable_2023} or duel channel GNNs~\cite{gao_addressing_2023} to provide graph national causing abnormality and augment neighbors based on similarity. 
As for deep-learning methods, approaches are applied in detection models such as meta-learning~\cite{ding_Fewshot_2021}, contrastive learning~\cite{duan_Graph_2023a,wang_CrossDomain_2023}, and active learning~\cite{dong_Active_2023}. The biggest challenge of this method is applying these deep-learning methods to instruct GAD. However, existing GAD methods often assume that clean and correct labels are available despite the expensive cost of confident annotations. Inspired by hybrid methods, this paper designs an effective detection model combined with reinforcement learning to address the challenge of noisy labels. 

\subsection{Noisy Label Learning on Graph}
The impact of noisy labels, including incomplete, inexact, and inaccurate labels in graphs, is relatively underexplored~\cite{xia_Robust_2021,du_Noiserobust_2021,dai_NRGNN_2021}. Graphs would render noisy information to truth labels and lead to poor detection performance. Current research endeavors meta-learning and augmentation to improve robustness. Meta\-GIN~\cite{ding_Robust_2022} obtains noise-reduced representations by interpolation and utilizes meta-learning to reduce label noise. Similarly, LPM~\cite{xia_Robust_2021} introduces meta-learning to optimize the label propagation, thereby reducing the negative effects of noisy information. RTGNN~\cite{qian_Robust_2023} and NRGNN~\cite{dai_NRGNN_2021} link unlabeled nodes with labeled nodes according to high similarity to augment graphs and predict pseudo labels for unlabeled nodes. Additionally, PIGNN~\cite{du_Noiserobust_2021} utilizes pair interactions, and D2PT~\cite{liu_learning_2023} proposes dual channels of the initial graph to augment information to learn representations. 
However, the above methods are primarily designed for node classification tasks and may achieve sub-optimal performance for anomaly detection due to imbalanced distribution. Hence, we formulate the research problem of GAD with noisy labels that effectively leverage the available imperfect labels for effective anomaly detection. 

\subsection{Graph Reinforcement Learning}
Recent research has revolved around integrating Deep Reinforcement Learning (DRL) with GNN due to their complementary strengths in various tasks~\cite{nie_reinforcement_2023,munikoti_challenges_2022}, such as node classification with generalization. GraphMixup~\cite{wu_graphmixup_2023} and GraphSR~\cite{zhou_graphsr_2023} both employ reinforcement learning to augment minority classes by generating edges for unlabeled nodes, effectively addressing imbalanced data. Moreover, some research has explored the use of Reinforcement Learning for adversarial attacks in graph~\cite{sun_adversarial_2020, dai_Adversarial_2018} that employ generate virtual edges or detect attacked edges. However, designing effective reward functions remains challenging. Graph reinforcement learning is also utilized in tasks like explanation~\cite{wang_reinforced_2023} and sparsification~\cite{wickman_generic_2022}, identifying significant edges or prunes irrelevant ones. Furthermore, anomaly detection tasks based on the assumption of clean labels leverage RL to select and filter neighborhood information propagation in GNN~\cite{dou_enhancing_2020, ding_Anomalyresistant_2021, huang_AUCoriented_2022}, thereby simplifying anomaly node pattern learning. However, they consider all nodes as targets for information filtering, which poses a large search space for RL. These approaches are unsuitable for situations where noisy labels exist, especially in the case of imbalanced data. Since DRL has not yet been applied to address the issue of noisy labels, we tackle noisy label influence by selecting edges to reconstruct graph structure based on the policy-based method.

\section{Conclusion}
In this paper, we propose a \textit{policy-in-the-loop} framework, REGAD, for GAD tasks with noisy labels. Our research focuses on pruning edges to address noisy labels' ripple effects through the graph structure. REGAD leverages a tailored policy network to refine information propagation step by step and search for a strategy to prune proper edges with high quality. Moreover, while noisy labels lead to weak supervision for cutting, we design a novel reward relying on detection performance enhancement which is computed by comparing confident pseudo labels rather than noisy ground-truths. 
The base detector and edge pruner feedback important information mutually and consist of the complete loop. Experiments show that REGAD generally outperforms baseline methods on three real-world datasets. 

\newpage
\bibliographystyle{IEEEtran}
\bibliography{regad}


\end{document}